\documentclass[a4paper,fleqn]{cas-dc}
\usepackage[numbers,sort&compress,longnamesfirst]{natbib}
\usepackage{subfigure}
\usepackage{amssymb}
\usepackage{gensymb}
\usepackage{color, xcolor}
\usepackage{hyperref}
\usepackage{graphicx}
\usepackage{amsmath}

\def\tsc#1{\csdef{#1}{\textsc{\lowercase{#1}}\xspace}}
\tsc{WGM}
\tsc{QE}

\begin{document}
\let\WriteBookmarks\relax
\def\floatpagepagefraction{1}
\def\textpagefraction{.001}

\setlength\abovedisplayskip{1pt}
\setlength\belowdisplayskip{1pt}

\shorttitle{GaussianCAD: Robust Self-Supervised CAD Reconstruction from Three Orthographic Views Using 3D Gaussian Splatting}    

\shortauthors{Zheng Zhou et~al.}  

\title [mode = title]{GaussianCAD: Robust Self-Supervised CAD Reconstruction from Three Orthographic Views Using 3D Gaussian Splatting}  

\author[label1]{Zheng Zhou}[type=editor,
                        auid=000,
                        bioid=1
]\fnref{co-first}
\ead{15827718@qq.com}

\author[label1]{Zhe Li}[type=editor,
                        auid=000,
                        bioid=1
]\fnref{co-first}
\ead{lzlomy@163.com}

\author[label2]{Bo Yu}[type=editor,
                                    auid=000,
                                    bioid=1
]\fnref{co-first}
\ead{a7858833@bupt.edu.cn}

\author[label1]{Lina Hu}[type=editor,
                           auid=000,
                           bioid=1
]
\ead{32724886@qq.com}

\author[label1]{Liang Dong}[type=editor,
                             auid=000,
                             bioid=1
]
\ead{80993231@qq.com}

\author[label2]{Zijian Yang}[type=editor,
                              auid=000,
                              bioid=1
]
\ead{zijianyoung@bupt.edu.cn}

\author[label2]{Xiaoli Liu}[type=editor,
                             auid=000,
                             bioid=1
]
\ead{liuxiaoli134@bupt.edu.cn}

\author[label1]{Ning Xu}[type=editor,
                           auid=000,
                           bioid=1
]
\ead{85636333@qq.com}

\author[label1]{Ziwei Wang}[type=editor,
                             auid=000,
                             bioid=1
]
\ead{zwexcelwangwang@163.com}

\author[label2]{Yonghao Dang}[type=editor,
                                auid=000,
                                bioid=1
]
\ead{dyh2018@bupt.edu.cn}

\author[label2]{Jianqin Yin}[type=editor,
                                         auid=000,
                                         bioid=1
]
\corref{cor}
\ead{jqyin@bupt.edu.cn}

\cortext[cor]{Corresponding author}
\fntext[co-first]{Equal Contribution}

\address[label1]{State Grid Hubei Electric Power Co., Ltd. Information and Communication Company, Hubei, China.}
\address[label2]{School of Intelligent Engineering and Automation, Beijing University of Posts and Telecommunications, Beijing, China.}

\begin{abstract}
The automatic reconstruction of 3D computer-aided design (CAD) models from CAD sketches has recently gained significant attention in the computer vision community. Most existing methods, however, rely on vector CAD sketches and 3D ground truth for supervision, which are often difficult to be obtained in industrial applications and are sensitive to noise inputs. We propose viewing CAD reconstruction as a specific instance of sparse-view 3D reconstruction to overcome these limitations. While this reformulation offers a promising perspective, existing 3D reconstruction methods typically require natural images and corresponding camera poses as inputs, which introduces two major significant challenges: (1) \textbf{modality discrepancy} between CAD sketches and natural images, and (2) \textbf{difficulty of accurate camera pose estimation} for CAD sketches. To solve these issues, we first transform the CAD sketches into representations resembling natural images and extract corresponding masks. Next, we manually calculate the camera poses for the orthographic views to ensure accurate alignment within the 3D coordinate system. Finally, we employ a customized sparse-view 3D reconstruction method to achieve high-quality reconstructions from aligned orthographic views. By leveraging raster CAD sketches for self-supervision, our approach eliminates the reliance on vector CAD sketches and 3D ground truth. Experiments on the Sub-Fusion360 dataset demonstrate that our proposed method significantly outperforms previous approaches in CAD reconstruction performance and exhibits strong robustness to noisy inputs.
\end{abstract}


\maketitle

\section{Introduction}

\begin{figure*}[Figure 1]
    \centering
    \subfigure{
        \includegraphics[width=1\textwidth]{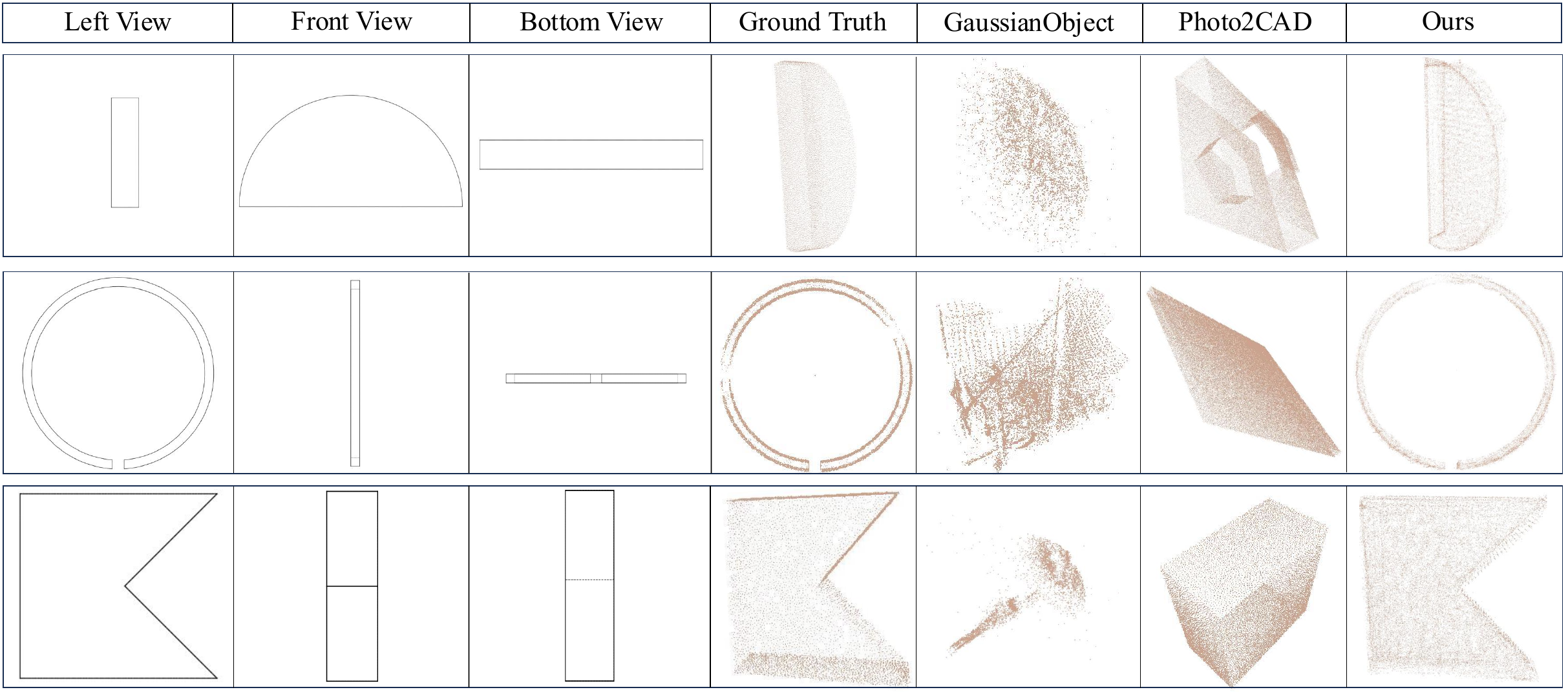}
    }
\begin{flushleft}
\caption{Reconstruction results obtained with GaussianCAD. The framework is capable of reconstructing high-quality 3D CAD models from raster CAD sketches containing three orthographic views based on 3D Gaussian Splatting. GaussianCAD demonstrating superior  performance over previous methods across diverse CAD sketches.}\label{Figure 1}
\end{flushleft}
\end{figure*}

The automatic reconstruction of 3D computer-aided design (CAD) models from CAD sketches has been a challenging problem in computer vision, computer graphics, and engineering design. This task is crucial in engineering, manufacturing, and architecture, as it dramatically enhances automation in design workflows. In this paper, we specifically focus on reconstructing 3D CAD models from raster CAD sketches containing three orthographic views.

Unlike natural images, CAD sketches consist solely of contours and geometric features, lacking materials and textures. Interpreting these sketches requires a deep understanding of engineering principles. To overcome these challenges, traditional CAD reconstruction methods \cite{yan1994efficient, furferi20102d, gorgani20193d, gong2006reconstruction, varley2010new, zhang2023automatic} typically follow a multi-stage rule-based process in which 3D vertices, edges, faces, and blocks are progressively constructed based on the results of previous steps. The key strength of this framework lies in its reliance on vector CAD sketches, which ensures high precision in matching the input views. However, this approach is highly susceptible to minor errors and noise inherent in the vector sketches, such as a pixel-level shift of an endpoint, resulting in significantly degraded performance.

Several approaches \cite{jones2020shapeassembly, hu2023plankassembly} recently have utilized end-to-end deep networks to reconstruct 3D CAD models from raster CAD sketches containing three orthographic views. These methods rely on a domain-specific language (DSL), a specialized language that describes CAD operations and geometry. They have achieved impressive reconstruction results for 3D CAD models represented by their DSL and have demonstrated high robustness to noisy inputs thanks to deep networks. However, the generalization of this approach is limited by the DSL itself. They can only reconstruct models defined using their DSL, and each type of model requires the creation of a new DSL.

In this paper, we introduce a novel CAD reconstruction method, named GaussianCAD, to overcome the reliance on vector CAD sketches and 3D ground truth. Our key innovation is that CAD reconstruction is a special case of sparse-view reconstruction. Sparse-view reconstruction is the task of reconstructing a 3D model from a limited number of input views, which is inherently ill-posed due to insufficient data. However, unlike natural images, which may not ensure a unique 3D model even with a 360-degree view, CAD sketches containing three orthographic views can uniquely determine a 3D CAD model without ambiguity. This capability alleviates the ill-posed problems of sparse-view reconstruction. Consequently, sparse-view reconstruction methods hold considerable promise for application in CAD reconstruction tasks. 

However, directly applying existing 3D reconstruction methods is problematic due to two key challenges: \textbf{bridging the modality discrepancy} and \textbf{estimating camera poses} for CAD sketches. First, these methods are designed for natural images, while CAD sketches consist only of lines, making it hard to adapt the underlying models. Second, the lack of rich visual features in CAD sketches makes it difficult to apply camera pose estimators developed for natural images. 

To solve these issues, we propose a three-stage process. First, the multi-view raster CAD sketches are fed into the Sketch Filtering module to detect edges, including dashed lines that represent invisible lines, which are subsequently removed. Second, the foreground of each view is determined based on the outermost edges detected in the first stage and then colorized. Third, the world-to-camera (w2c) extrinsic and camera intrinsic matrices are calculated to position the orthographic views within 3D space. After these steps, we employ an off-the-shelf sparse-view 3D reconstruction method based on 3D Gaussian Splatting (3DGS) \cite{kerbl20233d} to generate the 3D CAD model from the positioned views. Our approach operates within a self-supervised learning framework, eliminating the need for 3D ground truth. Additionally, the reconstruction process leverages deep learning-based 3D reconstruction techniques that are robust to noisy inputs. Furthermore, by utilizing 3D Gaussian Splatting (3DGS) as our 3D representation, the reconstructed models can be easily visualized and used for further applications.

To evaluate our method, we conducted experiments on Sub-Fusion360 \cite{zhang2023automatic} based on Fusion360 \cite{willis2021fusion} datasets, which contain thousands of CAD sketches and their corresponding 3D models. Our method demonstrates promising performance on the dataset, as shown in Fig. \ref{Figure 1}.

The contributions of this paper are summarized as follows:
\begin{itemize}
  \item [1)] 
    We propose a novel CAD reconstruction framework named GaussianCAD, the first to leverage existing 3D reconstruction methods for reconstructing 3D CAD models from raster CAD sketches containing three orthographic views. By operating within a self-supervised learning framework, our approach eliminates the reliance on vector CAD sketches and 3D ground truth.
    
  \item [2)]
    We present an effective module that transforms raster CAD sketches into representations resembling natural images and accurately estimates their corresponding camera poses. This effectively addresses bridging modality discrepancy and resolving camera pose estimation for CAD sketches.

  \item [3)]
    Experiments demonstrate that GaussianCAD achieves state-of-the-art (SOTA) results and excels with noisy inputs across various CAD sketches.
\end{itemize}

\section{Related Work}
\subsection{Traditional methods}
Various methods have been proposed for reconstructing 3D CAD models from CAD sketches. Early 3D reconstruction approaches \cite{yan1994efficient, furferi20102d} start with automatically converting CAD sketches to 3D wireframes. \cite{gorgani20193d} leveraging vertex labeling and fuzzy theory for reconstructing wireframes and surfaces. However, these methods did not consider curved edges. \cite{gong2006reconstruction} initiate a decision-tree-based method for reconstructing 3D wireframes of curved objects, which is limited by high algorithmic complexity. Varley et al. \cite{varley2010new} and Zakharov et al. \cite{zhang2023automatic} explored algorithms for identifying surfaces in wireframes, but they faced challenges with coplanar and nested loops. Although these methods have achieved high F-scores for well-reconstructed models on a public dataset containing several thousand 3D shapes, they are not robust enough to handle noisy input because they rely on precise alignment of vertices, edges, faces, and blocks based on exact coordinates in vector CAD sketches. Instead, our method learns an explicit scene representation by employing deep learning techniques that effectively handle noise and variations in input raster CAD sketches, thereby making the reconstruction process more robust.

\subsection{Deep Learning methods}
Integrating deep learning methods into 3D CAD modeling has recently garnered significant scholarly interest, encompassing transformer-based and convolutional neural network (CNN) approaches. For instance, Li et al. \cite{li2022free2cad, li2020sketch2cad} developed a technique that decomposes sketches into sequential CAD operations, utilizing a sequence-to-sequence neural network to segment clusters of lines effectively. Similarly, Wu et al. \cite{wu2021deepcad} delineated CAD models as ordered sequences of CAD commands, which are subsequently transformed into B-Rep models. This framework introduces a novel paradigm for automating and optimizing the CAD modeling process. These command-driven strategies have demonstrated remarkable efficacy in the realm of CAD modeling. Nonetheless, to our knowledge, no existing method has yet employed orthographic sketches as input.

Recent methods have investigated the ability of domain-specific language (DSL) to represent precise geometric CAD models. For example, ShapeAssembly \cite{jones2020shapeassembly} is a high-level programmatic framework that generates 3D shapes through the recursive composition of geometric primitives constrained by DSL. Similarly, PlankAssembly \cite{hu2023plankassembly} introduces a novel network design that learns shape programs to assemble planks into 3D cabinet models using a DSL. Although they produce promising results and exhibit excellent robustness to noisy inputs, it's also a double-edged sword because the rigid DSL constraints limit their flexibility, making them less applicable. Moreover, these deep learning-based approaches rely on ground truth 3D models for supervision, which is often difficult to obtain in industrial settings. In contrast, our method utilizes an off-the-shelf sparse-view 3D reconstruction method, which does not require 3D ground truth for supervision and offers greater flexibility and applicability.

\subsection{Sparse-view reconstruction methods} 
As an emerging 3D representation method, 3D Gaussian Splatting (3DGS) \cite{kerbl20233d} has been replacing Neural Radiance Field (NeRF) \cite{mildenhall2021nerf} in many scenarios, such as sparse-view reconstruction. FSGS \cite{zhu2025fsgs} was the pioneering method utilizing 3DGS to reconstruct 3D scenes from sparse-view inputs. It begins by initializing a sparse set of Gaussians using Structure-from-Motion (SfM) \cite{schonberger2016structure} techniques and refines these Gaussians through unpooling existing ones. GaussianObject \cite{yang2024gaussianobject} improves upon this by initializing Gaussians using a visual hull and refining them by fine-tuning a pre-trained ControlNet \cite{zhang2023adding} to repair degraded rendered images—degradations caused by adding noise to Gaussian properties. The required number of input views is drastically reduced to just four, representing a significant improvement over FSGS, which requires more than 20 views. In our paper, we employ GaussianObject as our sparse-view CAD reconstruction method, adapting it to our CAD sketches by omitting the fine-tuning process on Stable Diffusion \cite{rombach2022high} and excluding depth loss during training. Although GaussianObject initially demonstrates only four-view reconstruction, our modified approach achieves satisfactory results on three orthographic views.

\begin{figure*}[Figure 2]
    \centering
    \subfigure{
        \includegraphics[width=1\textwidth]{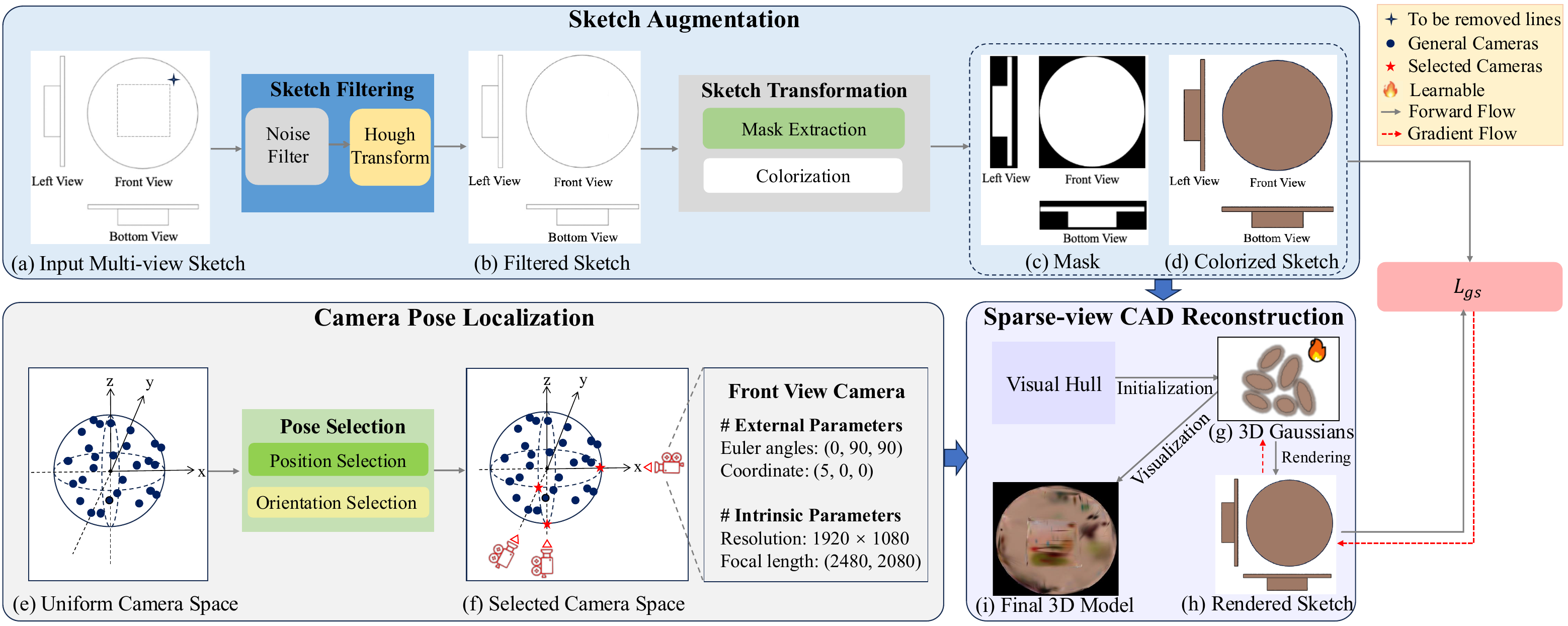}
    }
\begin{flushleft}
\caption{Overview of GaussianCAD. During \textbf{Sketch Augmentation}, a raster CAD sketch containing three orthographic views is provided as input. For each view, we filter out the noise and remove dashed lines, then extract the mask and colorize the corresponding foreground, ultimately obtaining a natural image–like CAD sketch. In the \textbf{Camera Pose Localization} stage, we determine the camera poses for the three views based on their relative positional priors to achieve precise alignment in 3D space and set uniform intrinsic camera parameters. During \textbf{Sparse-view CAD Reconstruction}, we initialize the 3D Gaussians by constructing the visual hull using the masked views and corresponding camera parameters, which are optimized with $\mathcal{L}_{gs}$. Finally, we obtain a 3D CAD model that matches the input multi-view sketch.}\label{Figure 2}
\end{flushleft}
\end{figure*}

\section{Method}
Given a raster CAD sketch consisting of three orthographic views with predetermined relative positions and orientation priors that describe a 3D CAD model, our goal is to reconstruct a high-quality 3D CAD model that remains consistent with the provided sketch. In this paper, we introduce a tailored process pipeline to adapt CAD sketches for effective 3D reconstruction (Sec. \ref{section:sketch filtering}), as illustrated in Fig. \ref{Figure 2}(a)-(d). Initially, we filter noise in the CAD sketches and eliminate dashed lines to retain only visible edges. Subsequently, precise masks are extracted to distinguish the foreground and background, and the sketches are colorized to provide meaningful color information necessary for supervising the reconstruction process. We then manually calculate the camera poses for orthographic views, ensuring accurate alignment of the views within the 3D coordinate system, which is critical for reliable reconstruction results (Sec. \ref{section:camera pose localization}), as illustrated in Fig. \ref{Figure 2}(e)-(f). Finally, we employ a sparse-view reconstruction method called GaussianObject, with tailored modifications, to achieve high-quality reconstruction using a single input CAD sketch with only three orthographic views (Sec. \ref{section:sparse view reconstruction}), as illustrated in Fig. \ref{Figure 2}(g)-(i).

\subsection{Preliminary}
\label{section:preliminary}
As demonstrated in \cite{spong2020robot}, Euler angles and rotation matrices are fundamental tools for representing three-dimensional orientations. This section introduces these concepts and camera parameters to provide the background for the camera pose localization presented in the methodology (Sec. \ref{section:camera pose localization}).

\textbf{Euler Angles.} Euler angles provide a systematic framework for describing the orientation of a rigid body in three-dimensional space through a sequence of three elemental rotations about specified axes. Typically, these rotations are denoted as $\alpha$, $\beta$, and $\gamma$, corresponding to rotations about the $z$-, $y$-, and $x$-axes, respectively.

\textbf{Rotation Matrix.} A rotation matrix is a $3 \times 3$ orthogonal matrix with a determinant equal to one, describing a vector's rotation in three-dimensional space. The general form of a rotation matrix $R$ is:

\begin{equation}
\hspace{4em} R = 
\begin{bmatrix}
R_{11} & R_{12} & R_{13} \\
R_{21} & R_{22} & R_{23} \\
R_{31} & R_{32} & R_{33} \\
\end{bmatrix}
\end{equation}

\textbf{Euler Angles to Rotation Matrix.} The rotation matrix derived from Euler angles $\alpha$, $\beta$, and $\gamma$ is obtained by applying individual rotation matrices about the principal axes. Specifically, the combined rotation matrix $R$ is given by:

\begin{equation}
    \hspace{4em} R = R_z(\alpha) R_y(\beta) R_x(\gamma)
\end{equation}

Where rotation matrices $R_z(\alpha), R_y(\beta)$, and $R_x(\gamma)$ are defined individually as:

\begin{align}
\hspace{3em} R_z(\alpha) &= 
\begin{bmatrix}
\cos\alpha & -\sin\alpha & 0 \\
\sin\alpha & \cos\alpha  & 0 \\
0          & 0           & 1 \\
\end{bmatrix} \label{Rz} \\
\hspace{3em} R_y(\beta) &= 
\begin{bmatrix}
\cos\beta  & 0 & \sin\beta \\
0          & 1 & 0        \\
-\sin\beta & 0 & \cos\beta \\
\end{bmatrix} \label{Ry} \\
\hspace{3em} R_x(\gamma) &= 
\begin{bmatrix}
1 & 0           & 0          \\
0 & \cos\gamma  & -\sin\gamma \\
0 & \sin\gamma  & \cos\gamma  \\
\end{bmatrix} \label{Rx}
\end{align}

\textbf{Extrinsic Matrix}. An extrinsic matrix $T$ in three-dimensional space is a $4 \times 4$ matrix that combines rotation, translation, scaling, and shearing, expressing the destination camera's pose relative to the source camera. It is typically expressed as:

\begin{equation}
\hspace{3em} T = 
\begin{bmatrix}
R_{11} & R_{12} & R_{13} & t_x \\
R_{21} & R_{22} & R_{23} & t_y \\
R_{31} & R_{32} & R_{33} & t_z \\
0      & 0      & 0      & 1   \\
\end{bmatrix}
\end{equation}

Where:
\begin{itemize}
    \item $R_{ij}$ elements form the upper-left $3 \times 3$ rotation matrix.
    \item $t_x$, $t_y$, $t_z$ represent translations along the respective axes.
    \item The last row $(0 \ 0 \ 0 \ 1)$ ensures the matrix operates correctly in homogeneous coordinates.
\end{itemize}

\textbf{Intrinsic Matrix}. The intrinsic matrix $K$ is a $3 \times 3$ matrix that encapsulates the camera's internal parameters, such as focal length and principal point, defining the camera's internal coordinate system. It is typically expressed as:

\begin{equation}
\hspace{5em} K =
\begin{bmatrix}
f_x & 0   & c_x \\
0   & f_y & c_y \\
0   & 0   & 1   \\
\end{bmatrix}
\end{equation}

Where:
\begin{itemize}
\item  $f_x$ and $f_y$  represent the focal lengths in terms of pixels along the  $x$ - and  $y$ -axes, respectively.
\item  $c_x$ and $c_y$  denote the coordinates of the principal point, typically located at the center of the image.
\item The last row $(0 \ 0 \ 1)$ ensures the matrix operates correctly in homogeneous coordinates.
\end{itemize}

\subsection{Sketch Augmentation}
\label{section:sketch augmentation}
The \textbf{modality discrepancy} between CAD sketches and natural images is a significant challenge in applying existing 3D reconstruction methods to CAD models. This section outlines these differences and introduces two key processes to address the issue: Sketch Filtering and Sketch Transformation. 

\begin{figure}
    \centering
    \subfigure{
        \includegraphics[width=0.48\textwidth]{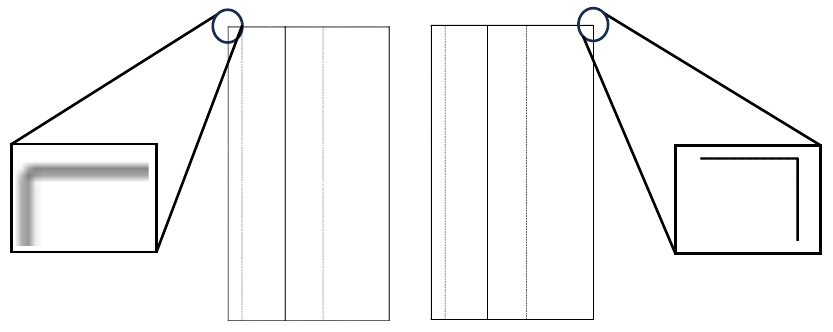}
    }
\begin{flushleft}
\caption{Visualization highlighting the quality differences between raster and vector CAD sketches. The raster sketch (left) exhibits significantly more noise than the vector sketch (right).}\label{Figure 3}
\end{flushleft}
\end{figure}
\textbf{Sketch Filtering.}
\label{section:sketch filtering}
Raster CAD sketches often incorporate artifacts and hidden lines, represented by dashed lines, as shown in Fig. \ref{Figure 3}. However, 3D reconstruction methods can only reconstruct objects from natural images, which capture visible surfaces. To address this issue, we first denoise the sketches. Then, using cleaner sketches, we apply the Canny edge detector \cite{canny1986computational} and the Hough Transform \cite{duda1972use} to detect and remove all dashed lines. Fig. \ref{Figure 2}(a)-(b) illustrates the Sketch Filtering process.

Specifically, to suppress these high-frequency noise components in the initial phase, we employ a median filter, which is more suitable than other filters because the lines in CAD sketches are often sparse and mainly consist of straight lines. We use a kernel size of $3 \times 3$, which needs to be carefully chosen to prevent thin lines from being filtered out due to adjacent large blank areas and the excessive size of larger filters. Then, the resulting cleaned sketch is binarized using an adaptive threshold to enhance contrast between dark lines and the light background. The resultant image exhibits reduced noise and minimal detail blurring, facilitating more robust feature extraction in subsequent stages.

After noise filtering, we apply the Canny edge detector and the Hough Transform to detect lines. Because the thickness, length, and pixel continuity of solid and dashed lines can be clearly distinguished in CAD sketches, we can detect and remove all dashed lines while preserving solid lines. First, we use the Canny edge detector to detect edges; then, we use the Hough Transform to extract geometric shapes from the edges, such as straight lines and circles. For each geometric shape detected, the corresponding pixels are extracted. The continuity of these pixels is then analyzed to segment the shapes into individual fragments or arc segments. Subsequently, the lengths and quantities of these segments or arcs are quantified. By comparing these measurements against predetermined thresholds, the shapes are classified as solid or dashed lines or circles. Ultimately, we obtain an edge map and remove all the dashed lines from the sketches.

\textbf{Sketch Transformation.}
\label{section:sketch transformation}
 Extracting precise masks from natural images is pivotal in 3D reconstruction methods. In general, masks of natural images can be attained by leveraging segmentation models, such as Segment Anything (SAM) \cite{kirillov2023segment}. Nonetheless, CAD sketches, characterized by their schematic representations and absence of intricate details, are not well-suited for direct use with off-the-shelf segmentation models. Instead, our method harnesses edges detected by the Canny edge detector to find the contours, thereby facilitating accurate mask generation indispensable for subsequent 3D reconstruction processes. Fig. \ref{Figure 2}(c) shows the extracted mask. 

To be precise, we first extract the contours using the Suzuki-Abe algorithm \cite{suzuki1985topological}, a contour detection method based on previously identified edges. We then create a binary mask with all pixel values initialized to zero (black). Then, we draw the identified contours onto this mask, filling them to produce a binary map where the areas within the contours are set to one (white). The mask is a definitive boundary delineation, distinguishing foreground and background for accurate 3D reconstruction.

Furthermore, to enhance the realism of the reconstructed model, we colorize the interior regions of the sketches corresponding to the masks with uniform colors. Users can arbitrarily adjust the sketch colors as desired, thereby influencing the appearance of the reconstructed 3D CAD model. Fig. \ref{Figure 2}(d) illustrates a colorized sketch.

\subsection{Camera Pose Localization}
\label{section:camera pose localization}

In addition to the modality discrepancy between CAD sketches and natural images, another critical challenge is \textbf{estimating camera poses} for CAD sketches. Natural images contain rich details and numerous matched feature points, which enable the utilization of image-matching methods, such as SfM \cite{schonberger2016structure} and DUSt3R \cite{wang2024dust3r} to estimate corresponding camera poses accurately. In contrast, CAD sketches typically lack such qualities, rendering these methods ineffective for pose estimation. As a result, the camera poses associated with CAD sketches must be manually configured.

Intrinsic and extrinsic parameters define a camera pose. For the intrinsic parameters, we employ a pinhole camera model with a resolution of \(1920 \times 1080\) pixels, with focal lengths of \(f_x = 2480\) pixels and \(f_y = 2080\) pixels, and the principal point located at \((960, 540)\) pixels for all cameras. The intrinsic matrix $K$ is thus defined as:

\begin{equation}
\hspace{3em} K =
\begin{bmatrix}
2480 & 0   & 960 \\
0   & 2080 & 540 \\
0   & 0   & 1   \
\end{bmatrix}
\end{equation}

The extrinsic parameters, which define the position and orientation of a camera, need to be calculated manually. For ease of verification, we begin with Blender's coordinate system and transform the cameras to the COLMAP's \cite{schonberger2016structure} coordinate system used by GaussianObject. In the Blender's coordinate system, the $x$-axis points to the camera's right, the $y$-axis points downwards, and the $z$-axis points towards the scene. To align the cameras with the desired poses, we set the Euler angles (in ZYX order) for the cameras to \((0, 90, 90)\), \((90, 0, 0)\), and \((0, 180, 90)\), and positions to \((0, 0, 5)\), \((0, -5, 0)\), \((0, 0, -5)\), which correspond to the front view, left view, and bottom view, respectively. We then calculate the corresponding rotation matrices according to the process detailed in Section \ref{section:preliminary}:
\begin{align}
\hspace{2em}
R_{\text{front}} &= R_z(0^\circ)\, R_y(90^\circ)\, R_x(90^\circ) \\
R_{\text{left}} &= R_z(90^\circ)\, R_y(0^\circ)\, R_x(0^\circ) \\
R_{\text{bottom}} &= R_z(0^\circ)\, R_y(180^\circ)\, R_x(90^\circ)
\end{align}

The extrinsic matrix $T$ for each camera is then constructed using the calculated rotation matrices and the specified translations:

\begin{align}
\hspace{3em}
T_{\text{front}} &= 
\begin{bmatrix}
R_{\text{front}} & \begin{bmatrix} 0 \\ 0 \\ 5 \end{bmatrix} \\
\begin{bmatrix} 0 & 0 & 0 \end{bmatrix} & 1
\end{bmatrix} \\
T_{\text{left}} &= 
\begin{bmatrix}
R_{\text{left}} & \begin{bmatrix} 0 \\ -5 \\ 0 \end{bmatrix} \\
\begin{bmatrix} 0 & 0 & 0 \end{bmatrix} & 1
\end{bmatrix} \\
T_{\text{bottom}} &= 
\begin{bmatrix}
R_{\text{bottom}} & \begin{bmatrix} 0 \\ 0 \\ -5 \end{bmatrix} \\
\begin{bmatrix} 0 & 0 & 0 \end{bmatrix} & 1
\end{bmatrix}
\end{align}

Since these camera poses are calculated based on their relative positions and orientations, there are no issues with estimation inaccuracies, which enhances the accuracy of the subsequent reconstruction process. Additionally, these poses can be applied to all sketches if they consist of the front, left, and bottom views.

\subsection{Sparse-view CAD Reconstruction}
\label{section:sparse view reconstruction}

We adapt the off-the-shelf sparse-view 3D reconstruction method, GaussianObject\cite{yang2024gaussianobject}, to our specific scenario to achieve high-quality CAD reconstruction. Specifically, apart from reconstructing from CAD sketch instead of natural images, we only use the Visual Hull for 3D Gaussians initialization and initial optimization from GaussianObject, as the subsequent steps in GaussianObject rely on Stable Diffusion \cite{rombach2022high} and LoRA \cite{hu2021lora} fine-tuned ControlNet \cite{zhang2023adding}. In general image reconstruction scenarios, this approach is practical. However, as Stable Diffusion lacks prior knowledge in the CAD domain, even this fine-tuning method fails to acquire sufficient information. Moreover, due to the absence of depth priors in CAD data, it is challenging to analyze CAD sketches directly to obtain accurate depth information. Therefore, we did not use depth loss in the initial optimization. The initial optimization involves two primary loss types: color loss and mask loss. The color loss is composed of an L1 loss and a D-SSIM loss derived from the 3D Gaussian Splatting:
\begin{align}
\hspace{6em}
   \mathcal{L}_{1} &= \Vert x - x^{\text{ref}} \Vert_1, \\
   \mathcal{L}_{D-SSIM} &= 1 - SSIM(x, x^{\text{ref}}),
\end{align}
where $x$ is the rendered sketch and $x^{ref}$ is the corresponding colorized reference sketch. 

For the mask loss, a binary cross-entropy (BCE) loss \cite{jadon2020survey} is utilized:
\begin{equation}
    \hspace{1em} \mathcal{L}_m = -(m^{ref} log\ m + (1-m^{ref})log(1-m))
\end{equation}
where $m$ denotes the mask of the reconstructed 3D CAD model, and $m^{ref}$ denotes the mask extracted from reference sketch.

The total loss function integrates these components as follows:
\begin{equation}
    \hspace{1em} \mathcal{L}_{gs} = (1 - \lambda_{SSIM})\mathcal{L}_1+\lambda_{SSIM}\mathcal{L}_{D-SSIM}+\lambda_{m}\mathcal{L}_m
\end{equation}
where $\lambda_{SSIM}$, $\lambda_{m}$, and $\lambda_{d}$ are weighting factors that balance the contributions of each term.

\section{Experiments}

\subsection{Dataset and Evaluation Metrics}

\textbf{Dataset}. We evaluate GaussianCAD \cite{yang2024gaussianobject} using Sub-Fusion360 \cite{zhang2023automatic} extracted from Fusion360 \cite{willis2021fusion} datasets, which comprises three orthographic views per CAD sketch. Shapes containing B-splines were excluded, and duplicate shapes were removed. The dataset consists of the original models, the generated CAD sketches, and the reconstructed boundary representation (B-Rep) models. Finally, it consists of 2981 samples. The CAD models contain between 3 and 447 edges and 3 to 151 faces. Since the dataset's sketches are vector-based, we captured screenshots and converted them to JPEG format at 1920 × 1080 pixels. Additionally, because our output consists of point clouds, we converted the B-Rep models into point clouds using FreeCAD \cite{li2022free2cad} for comparative analysis.

\textbf{Evaluation Metrics}. We first convert the wireframes into point clouds to facilitate comparison between the generated point clouds of CAD models and the wireframe ground truth. These are then normalized within a bounding box centered at the origin (0, 0, 0) with dimensions of 5 units for height, width, and length. Given that the coordinate systems of the generated point clouds do not align with the ground truth, we perform point cloud registration using the ground truth as the reference. We employ Chamfer Distance (CD), Hausdorff Distance (HD), and Earth Mover’s Distance (EMD) to assess the similarity of the point cloud shapes between the generated models and the ground truth.

\begin{itemize}
    \item \textbf{Chamfer Distance (CD)}: Measures the average minimal distance between points of two sets.
    \begin{equation}\scriptsize
        f_{CD}(A, B) = \frac{1}{|A|} \sum_{a \in A} \min_{b \in B} \|a - b\| 
        + \frac{1}{|B|} \sum_{b \in B} \min_{a \in A} \|b - a\|
    \end{equation}
    
    \item \textbf{Hausdorff Distance (HD)}: Identifies the maximum minimal distance between points of two sets.
    \begin{equation}\scriptsize
        f_{HD}(A, B) = \max\left\{\sup_{a \in A} \inf_{b \in B} \|a - b\|, 
        \sup_{b \in B} \inf_{a \in A} \|b - a\| \right\}
    \end{equation}
    
    \item \textbf{Earth Mover's Distance (EMD)}: Measures the minimum cost to transform one set into another.
    \begin{equation}\scriptsize
        f_{EMD}(A, B) = \min_{\phi: A \to B} \sum_{a \in A} \|a - \phi(a)\|
    \end{equation}
\end{itemize}

$A$ and $B$ refer to individual point clouds, respectively, while $a$ and $b$ denote individual points within these point clouds.

\subsection{Implementation Details}
\label{implementation details}
Our framework (Fig. \ref{Figure 2}) comprises three components. The first component, Sketch Augmentation (Sec. \ref{section:sketch augmentation}), is implemented using Probabilistic Hough Transform and Hough Circle Transform for straight line and circle detection.
The second component, Camera Pose Localization (Sec. \ref{section:camera pose localization}), determines the coordinates and Euler angles of cameras within Blender's coordinate system and computes their corresponding extrinsic matrices. Since the reconstruction model operates in COLMAP's coordinate system, we convert the camera poses from Blender's to COLMAP's. Additionally, we define the cameras' intrinsic parameters using Blender's default settings.
The third component (Sec. \ref{section:sparse view reconstruction}) employs GaussianObject for 3D reconstruction. We adhere to the original setup except for omitting steps after Visual Hull initiation and initial optimization. The 3DGS model is trained for 10k iterations during the optimization. Training a 3D CAD model takes approximately 2 minutes on a single NVIDIA RTX 4090 GPU.

\begin{figure*}[Figure 4]
    \centering
    \subfigure{
        \includegraphics[width=1\textwidth]{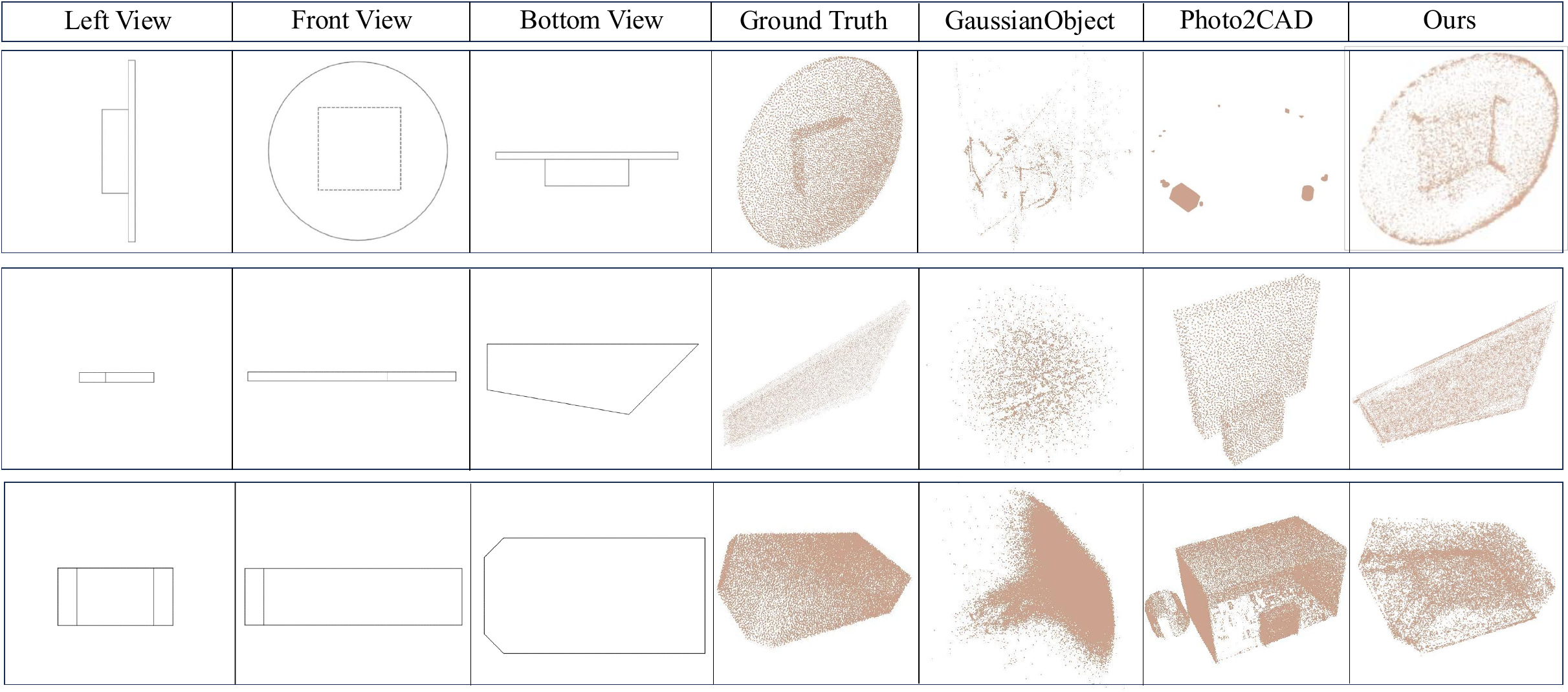}
    }
\begin{flushleft}
\caption{Qualitative examples from the Sub-Fusion360 dataset. GaussianObject fails to reconstruct any 3D CAD model, producing sparse artifacts and fragmented clusters. Photo2CAD reconstructs some 3D models, but they do not match the input CAD sketches, resulting in inaccurate outcomes.}\label{Figure 4}
\end{flushleft}
\end{figure*}

\subsection{Evaluation}
\emph{Baselines}. We evaluate GaussianCAD against several reconstruction baselines, including vanilla GaussianObject \cite{yang2024gaussianobject}, Photo2CAD \cite{harish2021photo2cad}, and OrthoRec \cite{zhang2023automatic}. Originally designed for reconstruction from natural images, the vanilla GaussianObject employs SAM \cite{kirillov2023segment} for mask extraction and Dust3R \cite{wang2024dust3r} for camera pose estimation. Photo2CAD takes raster sketches as input and outputs SCAD files; for comparison purposes, we convert them into STL format and visualize them using the Open3d library. All models
are trained using publicly released codes. OrthoRec is a traditional rule-based, multi-stage method that accepts vector sketches as input. Since no traditional methods are available as open-source implementations, we have exclusively reimplemented the latest method, OrthoRec, based on their published paper.

\emph{Sparse-view Reconstruction Performance}. To ensure a fair comparison, we evaluate our method alongside other approaches that reconstruct 3D models from raster images. Table \ref{table 1} shows the reconstruction fidelity of GaussianCAD compared to these methods. The experiments demonstrate that GaussianCAD achieves state-of-the-art (SOTA) results across all metrics. Fig. \ref{Figure 4} illustrates the reconstruction results of vanilla GaussianObject, Photo2CAD, and our method across different CAD sketches. GaussianCAD achieves significantly better quality and fidelity than the other methods. We find that vanilla GaussianObject completely collapses because the masks obtained from SAM and camera poses estimated by Dust3R are inaccurate; we will conduct ablation experiments on these components in Sec. \ref{section:ablation}. Regarding Photo2CAD, its performance is also notably poor because the design of its framework is limited to handling specific shapes, such as squares and circles. The result indicates that our method is more effective and generalizable than these alternatives.

\begin{table}[htbp]\normalsize 
\begin{center}
\renewcommand\arraystretch{1.4}
\renewcommand\tabcolsep{10pt}
\begin{flushleft}
\caption{Comparisons with existing methods. Smaller values for all metrics signify better performance. The best results are bolded.}\label{table 1}
\end{flushleft}
\begin{tabular}{cccc}
\hline
Method           & CD $\downarrow$   & HD $\downarrow$  & EMD $\downarrow$ \\ \hline
GaussianObject \cite{harish2021photo2cad} & 1.359  & 2.375 & 1.017 \\
Photo2CAD\cite{yang2024gaussianobject}     & 1.011  & 1.652 & 0.849 \\ 
Ours              & \textbf{0.307} & \textbf{0.589} & \textbf{0.353} \\ \hline
\end{tabular}
\end{center}
\end{table}

\begin{figure}
    \centering
    \subfigure{
        \includegraphics[width=0.47\textwidth]{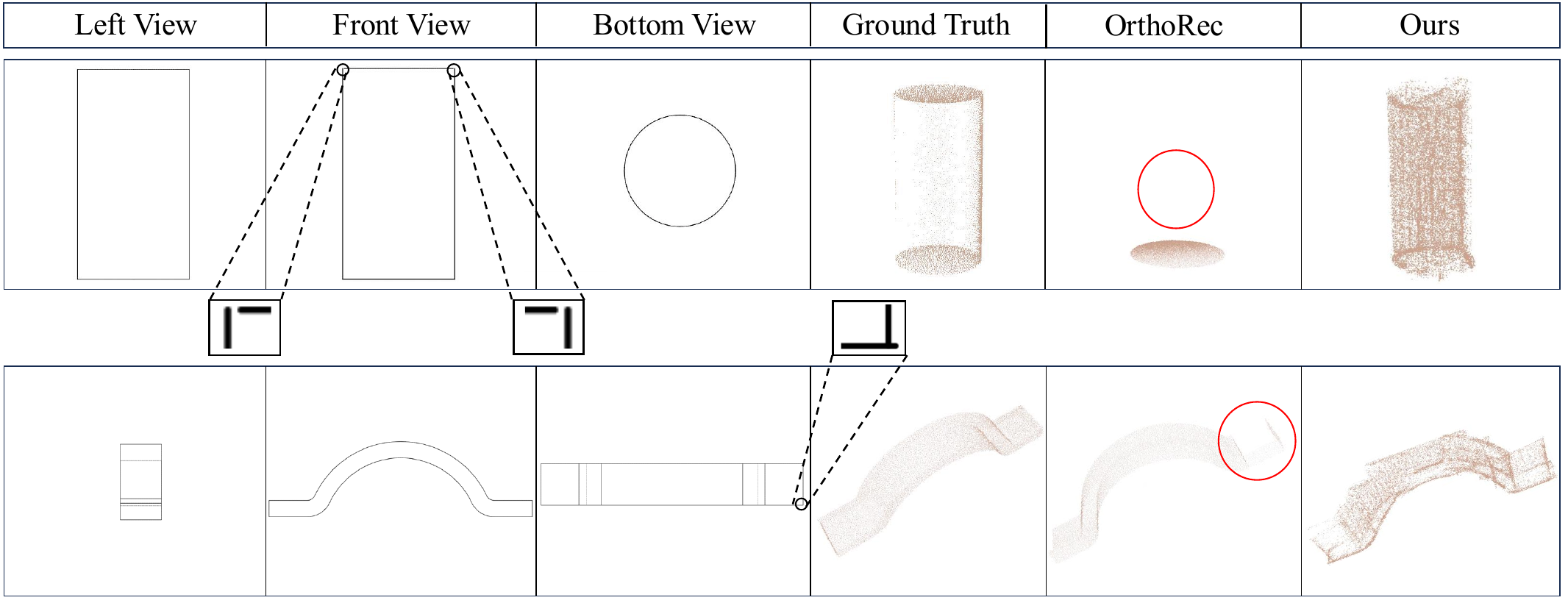}
    }
\begin{flushleft}
\caption{Robustness verification of our method. The results demonstrate that our approach remains robust under noisy input conditions, while the traditional method (OrthoRec) fails to reconstruct accurately even with minimal noise. Red circles highlight the missing faces.}\label{Figure 5}
\end{flushleft}
\end{figure}

\emph{Verification of Robustness to Input Noise}. Most traditional methods extract edges from CAD sketches in SVG format, which is a vector-based format. Their precise 2D edge feature matching relies on accurately extracted coordinates from SVG sketches. However, noise in SVG sketches can significantly impair the reconstruction process. We simulate common noise errors by extending or shortening edges to evaluate this. As shown in Fig. \ref{Figure 5}, even minimal noise—visible only after nearly fivefold magnification—causes OrthoRec \cite{zhang2023automatic} to miss one or more faces due to failed matches of vertices and edges across the views. In contrast, our method successfully reconstructs models from coarse raster CAD sketches that contain substantially more noise.

\subsection{Ablation Studies and Analysis}
\label{section:ablation}
In this section, we conduct ablation studies to validate the efficiency and essentiality of different modules.

\begin{figure}
    \centering
    \subfigure{
        \includegraphics[width=0.45\textwidth]{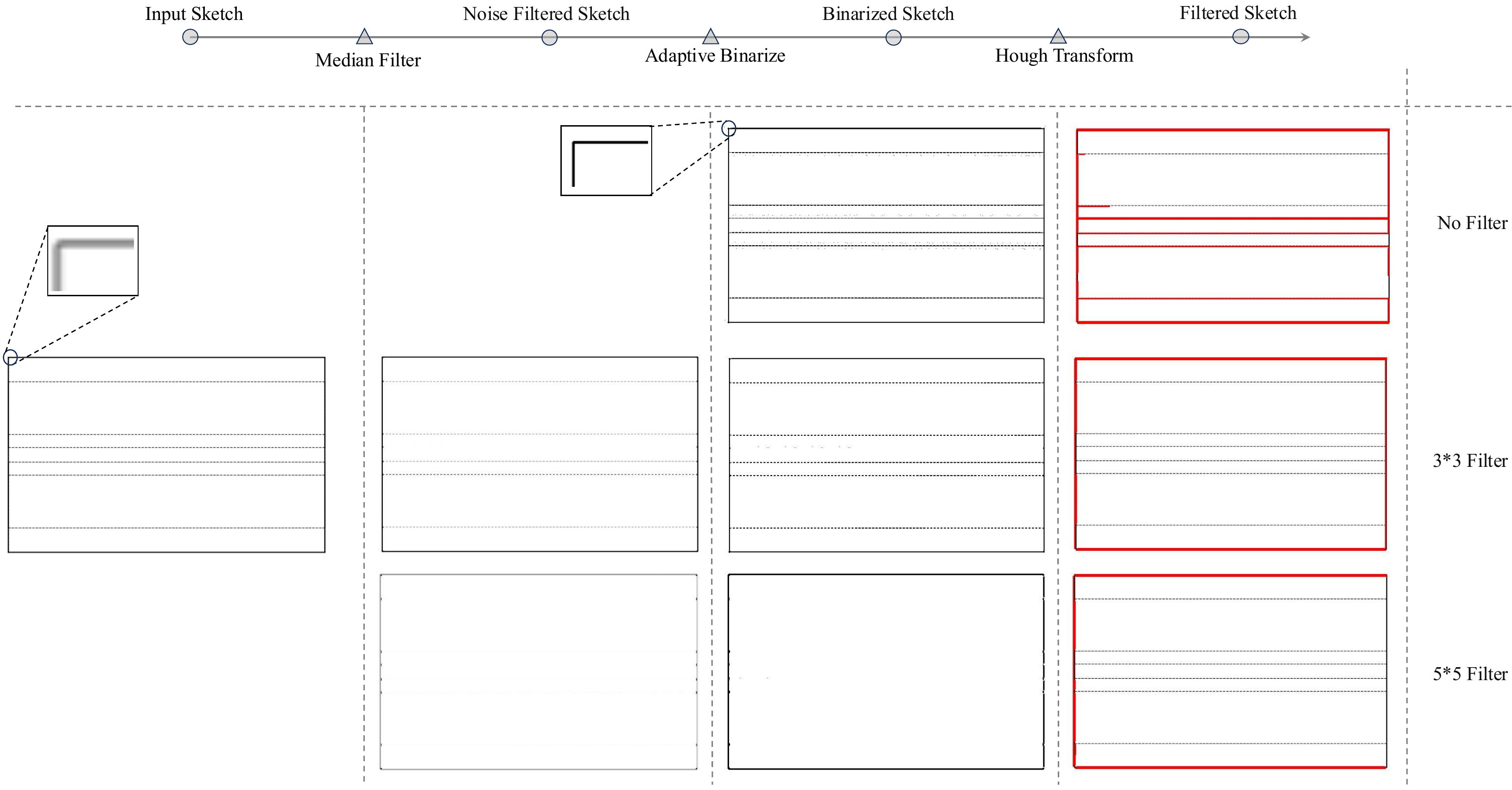}
    }
\begin{flushleft}
\caption{Ablation study on $3 \times 3$ median filter. This $3 \times 3$ filter provides the most robust denoising effect and enhances the performance of subsequent Hough Transform operations. Red lines indicate the extracted solid lines.}\label{Figure 6}
\end{flushleft}
\end{figure}

\emph{Impact of the kernel size of median filter}. The median filter plays a crucial role in the sketch Filter module. Since most lines in CAD sketches are thin and the light background occupies most of the sketch area, using a large kernel size causes dashed lines to vanish and diminishes solid lines, as shown in the bottom row of Fig. \ref{Figure 6}. If we do not use a filter, the Hough Transform cannot extract the solid lines, as illustrated in the top row of Fig. \ref{Figure 6}. In our experiments, a median filter with a kernel size of $3 \times 3$ is the most robust.

\begin{figure}
    \centering
    \subfigure{
        \includegraphics[width=0.47\textwidth]{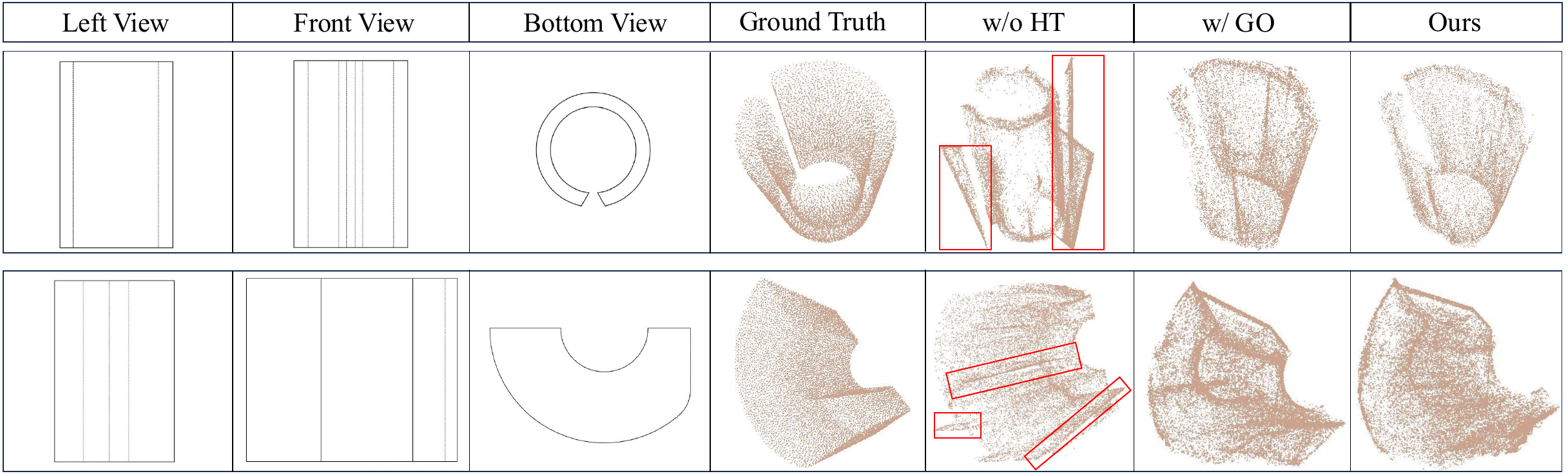}
    }
\begin{flushleft}
\caption{Ablation study on Hough Transform and Simplified GaussianObject. "HT" denotes Hough Transform. "GO" denotes the full GaussianObject. Red frames indicate the protrusions.}\label{Figure 7}
\end{flushleft}
\end{figure}

\emph{Impact of Hough Transform}.
We also perform an ablation study on the Hough Transform, which is employed to remove dashed lines. Because the training objective of 3D reconstruction methods is to minimize the difference between the reference image and the rendered image, the dashed lines will interfere with the estimation process of the depth information of the 3D reconstruction method, thereby affecting the generation of the depth map, resulting in uneven or abnormally protruding parts on the surface of the generated 3D model. As shown in Fig. \ref{Figure 7}, the results demonstrate the necessity of our sketch Filtering module.

\emph{Impact of Simplified GaussianObject}. To evaluate the effectiveness of our proposed modifications, We conducted an ablation study comparing our proposed Sketch Augmentation and Camera Pose Localization components, integrated into a simplified GaussianObject framework, with the full GaussianObject framework enhanced by these same components. As shown in Fig. \ref{Figure 7}, although we employed only a subset of GaussianObject, the reconstruction results for CAD sketches containing three orthographic views are satisfactory and not greatly different from the full GaussianObject. Additionally, the fine-tuning process of Diffusion is time-consuming; the original GaussianObject takes 35 minutes to run a reconstruction pipeline, while ours only takes 2 minutes. These results demonstrate our pipeline's effectiveness and efficiency.

\begin{figure}
    \centering
    \subfigure{
        \includegraphics[width=0.47\textwidth]{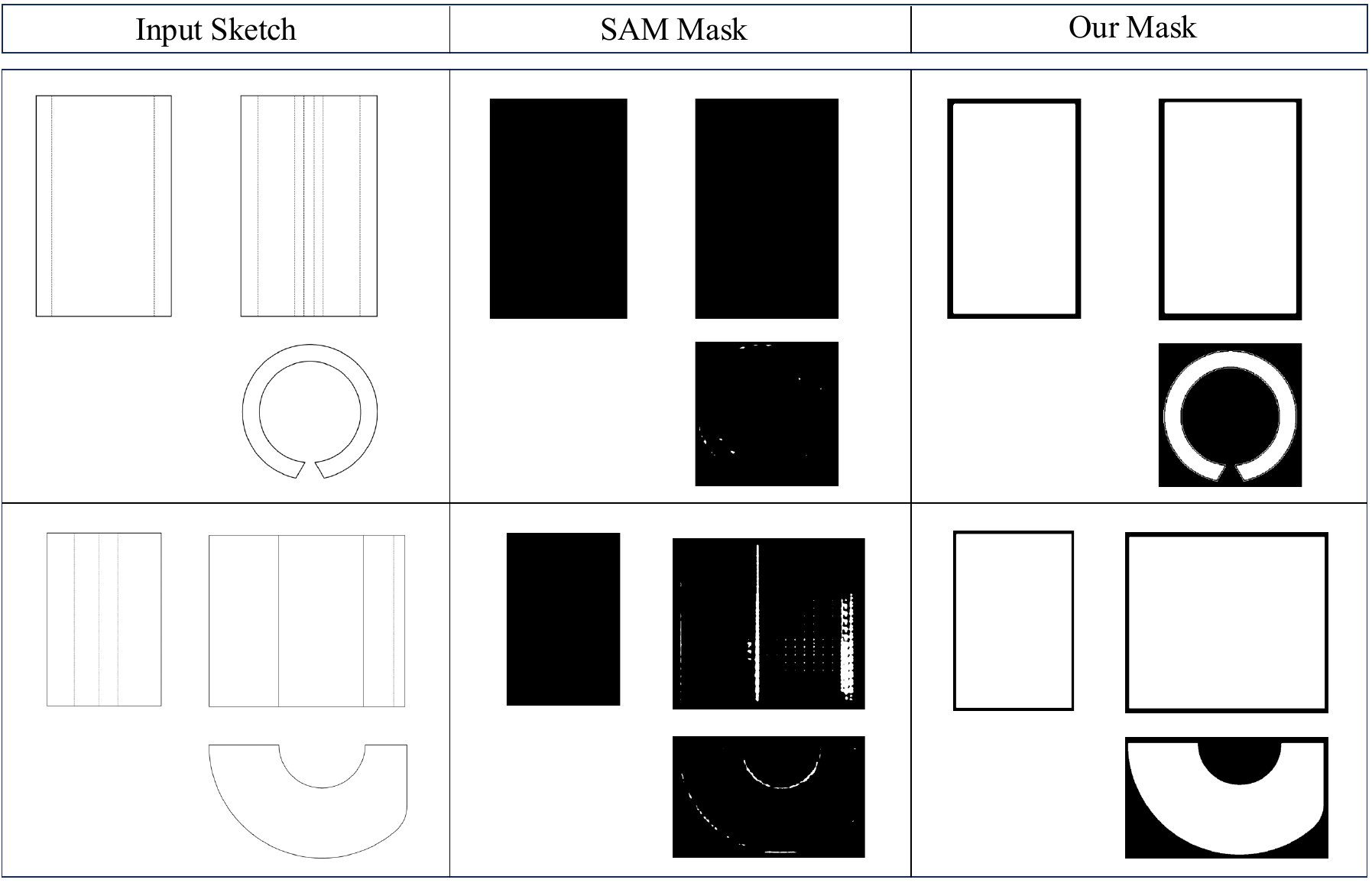}
    }
\begin{flushleft}
\caption{Ablation study on Mask Extraction. Compared to masks extracted by SAM, our method produces significantly more accurate masks. White areas represent the foreground, while black areas represent the background. A completely black image indicates a total failure in mask extraction.}\label{Figure 8}
\end{flushleft}
\end{figure}

\emph{Impact of Mask Extraction}.
Fig. \ref{Figure 8} illustrates the segmentation results of SAM used by GaussianObject. The results of SAM are significantly inaccurate, demonstrating that the large pre-trained segmentation model designed for natural images cannot detect edges and extract masks for CAD sketches. In contrast, our method can precisely detect all lines and extract the corresponding masks. The result confirms the effectiveness of our mask extraction module.

\begin{figure}
    \centering
    \subfigure{
        \includegraphics[width=0.47\textwidth]{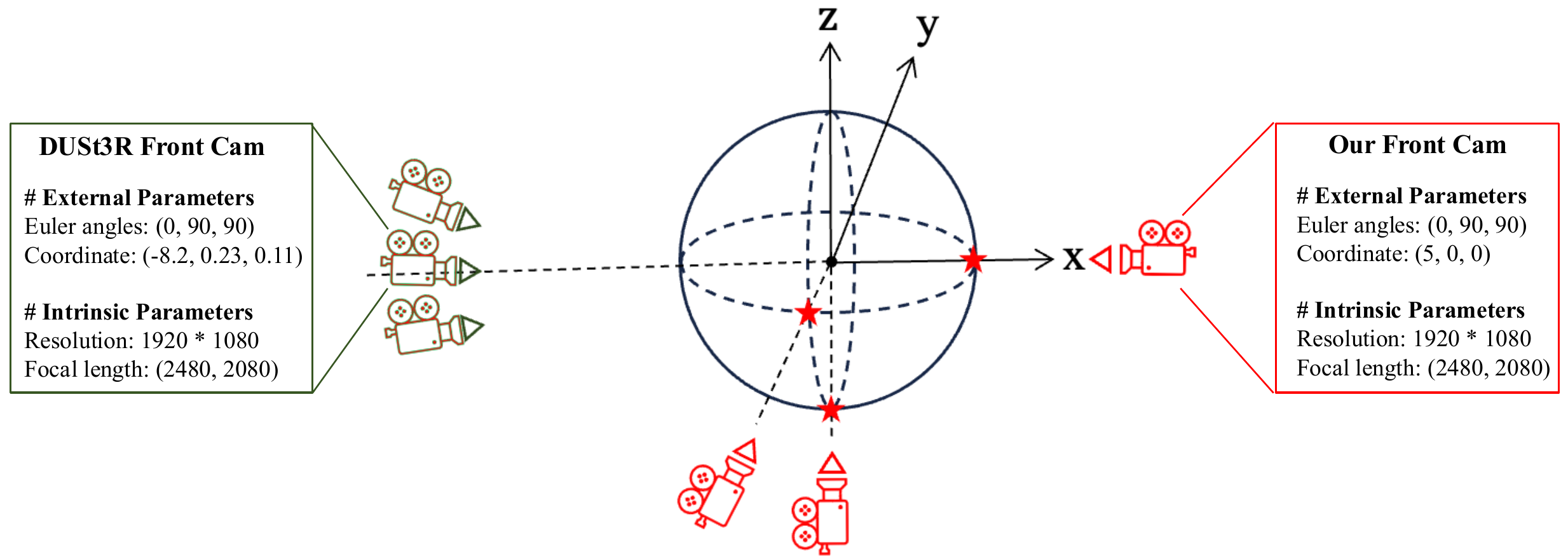}
    }
\begin{flushleft}
\caption{Ablation study on Camera Pose Selection. Compared to camera poses estimated by Dust3R, our method accurately determines the correct relative positions and orientations of the cameras.}\label{Figure 9}
\end{flushleft}
\end{figure}

\emph{Impact of Camera Poses Alignment}.
Fig. \ref{Figure 9} depicts the camera poses estimated by Dust3R, which are utilized in COLMAP-Free GaussianObject for natural images that lack precise camera pose information. These estimated camera poses are inaccurate. In contrast, our method localizes the camera poses for the corresponding views by leveraging relative positional priors, enabling unbiased localization and facilitating more accurate CAD reconstruction. Once localized, the camera poses can be directly applied to any other sketches, provided that they include front, left, and bottom views.

\begin{figure}
    \centering
    \subfigure{
        \includegraphics[width=0.47\textwidth]{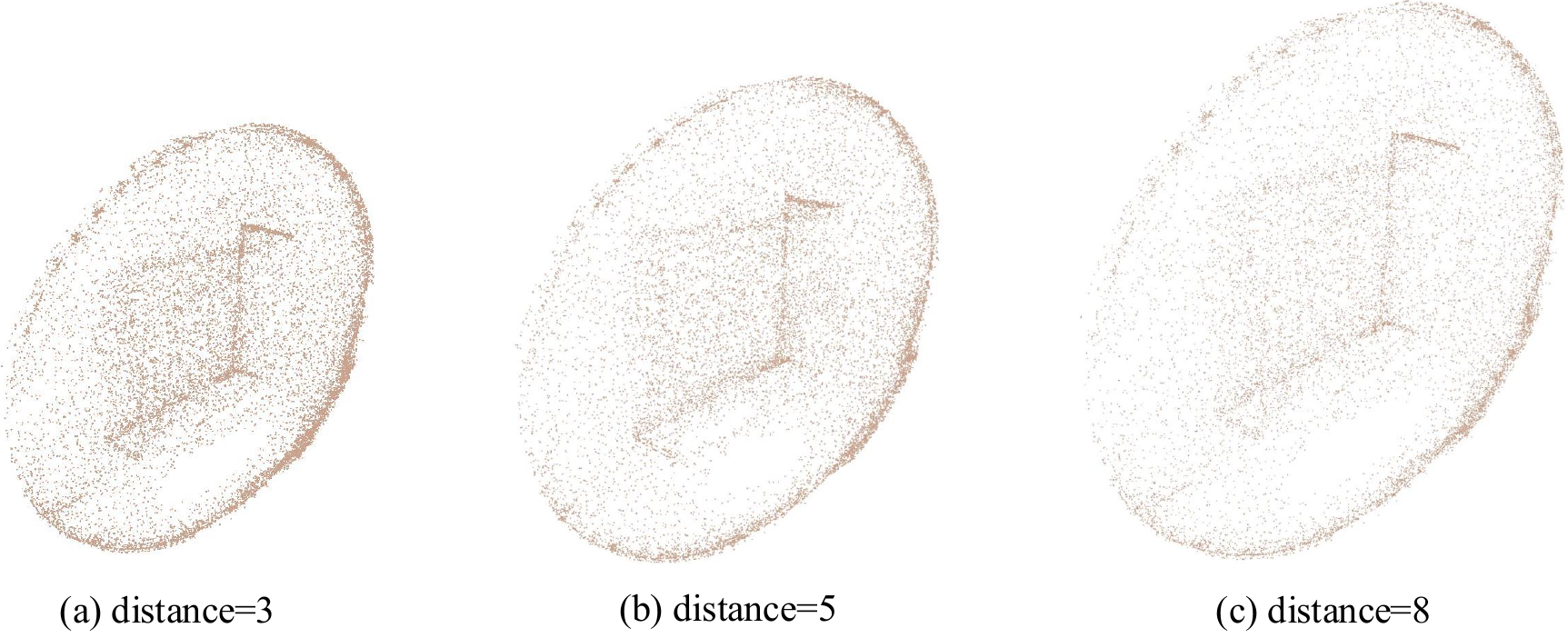}
    }
\begin{flushleft}
\caption{Ablation study on Camera Position Selection. The camera's distance from the origin only affects the size of the reconstructed models.}\label{Figure 10}
\end{flushleft}
\end{figure}
\emph{Impact of Absolute Camera Position Selection}. We do not ablate the relative positions and orientations of the cameras corresponding to the views because natural constraints inherently determine them. Instead, we ablate the selection of absolute camera positions. In our experiment, we assume that the target object is positioned at the origin and set the cameras at a distance of 5 units from the origin. This distance is comparable to the camera distances from objects in the MipNeRF360 \cite{barron2022mip} dataset. However, in practice, the distances can be set over a wide range, which only influences the size of the reconstructed model. The selection is solely restricted by the visual hull's initial size when the cameras' intrinsic parameters are fixed. If the distances are too far, points in the Visual Hull will not fall into the image plane, causing the reconstruction to fail. As shown in Fig. \ref{Figure 10}, the shape of the reconstructed models remains consistent under different distances.

\subsection{Limitations and Future work}
GaussianCAD exhibits outstanding performance in CAD reconstruction; however, areas warrant further exploration. For CAD sketches with very dense lines, noise significantly affects the effectiveness of the sketch filter, resulting in coarse or even incorrect mask extraction. Additionally, removing all dashed lines instead of utilizing them leaves the prior knowledge incomplete. Consequently, reconstruction may fail in cases where the three views are significantly different and complex. Analyzing the relationships between views and reasoning out opposing views may offer solutions to improve results.

Furthermore, while leveraging existing 3D reconstruction techniques yields satisfactory results, the 3D rendering methods that map three-dimensional objects onto two-dimensional representations—are not ideally suited for CAD sketches. This is because CAD sketches are drawn using orthogonal projection. Orthogonal projection maintains parallel lines but lacks a sense of depth. In contrast, real-world images use perspective projection to simulate human vision. In perspective projection, parallel lines appear to converge at vanishing points, creating a realistic perception of depth and dimension.

Nonetheless, the relatively small depth differences between front and back objects in CAD views, combined with precise multi-view consistency and alignment, mean that the differences do not significantly impact the results. Modifying the 3D rendering process is a promising and essential area for future work to enhance the precision and plausibility of CAD reconstruction.

\section{Conclusion}
In conclusion, we introduce GaussianCAD, a novel framework for reconstructing 3D CAD models from raster sketches containing three orthographic views. Our approach leverages advanced sparse-view reconstruction techniques based on 3DGS. Crucially, it alleviates the inherent ill-posed problems of sparse-view reconstruction by exploiting the unique correspondence between three orthographic views and a single CAD model. GaussianCAD consists of three key components: Sketch Augmentation, which prepares suitable inputs for the 3D reconstruction process; Camera Pose Localization, which ensures precise alignment of the orthographic views in 3D space; and Sparse-view CAD Reconstruction, which generates the final 3D model. Experimental results demonstrate that GaussianCAD achieves superior accuracy and robustness compared to existing methods.

\section{Declaration of competing interest}
The authors declare that they have no known competing financial interests or personal relationships that could have appeared to influence the work reported in this paper.

\section{Acknowledgments}

This work was supported by the State Grid Corporation Science and Technology Project Funding (Project Code: 5700-202422243A-1-1-ZN).

\bibliographystyle{model1-num-names}
\bibliography{cas-refs}
\end{document}